\documentclass[lettersize,journal]{IEEEtran}
\usepackage{amsmath,amsfonts}
\usepackage{algorithmic}
\usepackage{algorithm}
\usepackage{array}
\usepackage[caption=false,font=normalsize,labelfont=sf,textfont=sf]{subfig}
\usepackage{textcomp}
\usepackage{stfloats}
\usepackage{url}
\usepackage{multirow}
\usepackage{microtype}
\usepackage{tabularx}
\usepackage{colortbl}
\usepackage{booktabs}
\usepackage{verbatim}
\usepackage{graphicx}
\usepackage{cite}
\usepackage{adjustbox}
\usepackage{float}
\usepackage[table]{xcolor}
\usepackage[colorlinks,
            linkcolor=blue,
            anchorcolor=blue,
            citecolor=green, 
            urlcolor=blue]{hyperref}
\usepackage{orcidlink}

\newcolumntype{C}[1]{>{\centering\arraybackslash}p{#1}}

\begin{document}

\title{DET-GS: Depth- and Edge-Aware Regularization for High-Fidelity 3D Gaussian Splatting}

\author{Zexu Huang\orcidlink{0009-0006-6251-9694}, Min Xu\orcidlink{0000-0001-9581-8849},~\IEEEmembership{Member,~IEEE}, Stuart Perry\orcidlink{0000-0002-2794-3178},~\IEEEmembership{Senior Member,~IEEE}
\thanks{Zexu Huang, Min Xu and Stuart Perry are with the Perceptual Imaging Laboratory (PILab), School of Electrical and Data Engineering, Faculty of Engineering and Information Technology, University of Technology Sydney, Ultimo, NSW 2007, Australia (email: \href{mailto:zexu.huang@student.uts.edu.au}{zexu.huang@student.uts.edu.au}; \href{mailto:min.xu@uts.edu.au}{min.xu@uts.edu.au}; \href{mailto:stuart.perry@uts.edu.au}{stuart.perry@uts.edu.au})}
\thanks{Manuscript received August 6, 2025;}}



\maketitle

\begin{abstract}
3D Gaussian Splatting (3DGS) represents a significant advancement in the field of efficient and high-fidelity novel view synthesis. Despite recent progress, achieving accurate geometric reconstruction under sparse-view conditions remains a fundamental challenge. Existing methods often rely on non-local depth regularization, which fails to capture fine-grained structures and is highly sensitive to depth estimation noise. Furthermore, traditional smoothing methods neglect semantic boundaries and indiscriminately degrade essential edges and textures, consequently limiting the overall quality of reconstruction. In this work, we propose DET-GS, a unified depth and edge-aware regularization framework for 3D Gaussian Splatting. DET-GS introduces a hierarchical geometric depth supervision framework that adaptively enforces multi-level geometric consistency, significantly enhancing structural fidelity and robustness against depth estimation noise. To preserve scene boundaries, we design an edge-aware depth regularization guided by semantic masks derived from Canny edge detection. Furthermore, we introduce an RGB-guided edge-preserving Total Variation loss that selectively smooths homogeneous regions while rigorously retaining high-frequency details and textures. Extensive experiments demonstrate that DET-GS achieves substantial improvements in both geometric accuracy and visual fidelity, outperforming state-of-the-art (SOTA) methods on sparse-view novel view synthesis benchmarks. (The link to the code will be made available after publication)
\end{abstract}


\section{Introduction}
\IEEEPARstart{T}{he} field of 3D reconstruction has experienced significant advancements, notably enhancing innovative view synthesis and facilitating the display of photorealistic volumetric scenes. Neural radiance fields (NeRF)~\cite{mildenhall2021nerf} have achieved significant advancements in reconstructing photorealistic performances and precise 3D structures from sparse view sets~\cite{chen2021mvsnerf,yang2023freenerf,yu2021pixelnerf}. However, numerous sparse-input NeRF frameworks are constrained by slow computation speeds and high memory consumption. The identified limitations lead to substantial computational and temporal expense, thereby considerably hindering their application in practical applications. To address efficiency issues, some approaches integrate grid-based structures~\cite{muller2022instant,sun2022direct} to accelerate inference. However, these methods present inherent trade-offs: although they enhance rendering speed, they often lead to a substantial increase in training overhead or a reduction in the fidelity of the rendered images.

In recent work, 3D Gaussian Splatting (3DGS)~\cite{kerbl20233d} proposed an unstructured radiance field representation utilizing 3D Gaussian primitives, demonstrating exceptional performance in fast, high-quality, and cost-effective novel view synthesis when trained with densely sampled color images. Even under sparse view conditions, it successfully preserves detailed and well-defined regional characteristics. Nonetheless, 3DGS frequently encounters visual distortions when subjected to camera angles that were not included in its training or during proximity observations, which can be attributed to its restricted ability to reconstruct fine details. While several improvements~\cite{yu2024mip,lu2024scaffold,yang2024spec} have been proposed to enhance 3DGS through advanced smoothing filters~\cite{yu2024mip}, better initialization strategies~\cite{lu2024scaffold}, and more expressive appearance models~\cite{yang2024spec}, these approaches primarily focus on appearance preservation and do not directly address the underlying geometric accuracy. Depth information has recently emerged as a powerful geometric prior for Gaussian splatting frameworks. However, existing depth-regularized methods~\cite{10678533,yu2024lm,shen2025dof} typically impose supervision by enforcing pixel-to-pixel consistency between the predicted and reference depth maps over the entire image. Such supervision overlooks local geometric context, making it difficult to capture fine-grained structures and increasing sensitivity to monocular depth noise. Moreover, current regularization techniques~\cite{10678533,yu2024lm,li2024dngaussian,shen2025dof} apply uniform smoothing across the scene without considering semantic or geometric boundaries, resulting in blurred object contours and loss of structural details. While 3DGS models~\cite{kerbl20233d,lu2024scaffold,yu2024mip,li2024dngaussian,yang2024spec,shen2025dof} achieve high-quality color reconstructions, they lack image-space regularization that distinguishes flat regions from high-frequency areas.

\begin{figure*}[!t]
    \centering
    \includegraphics[width=0.85\textwidth]{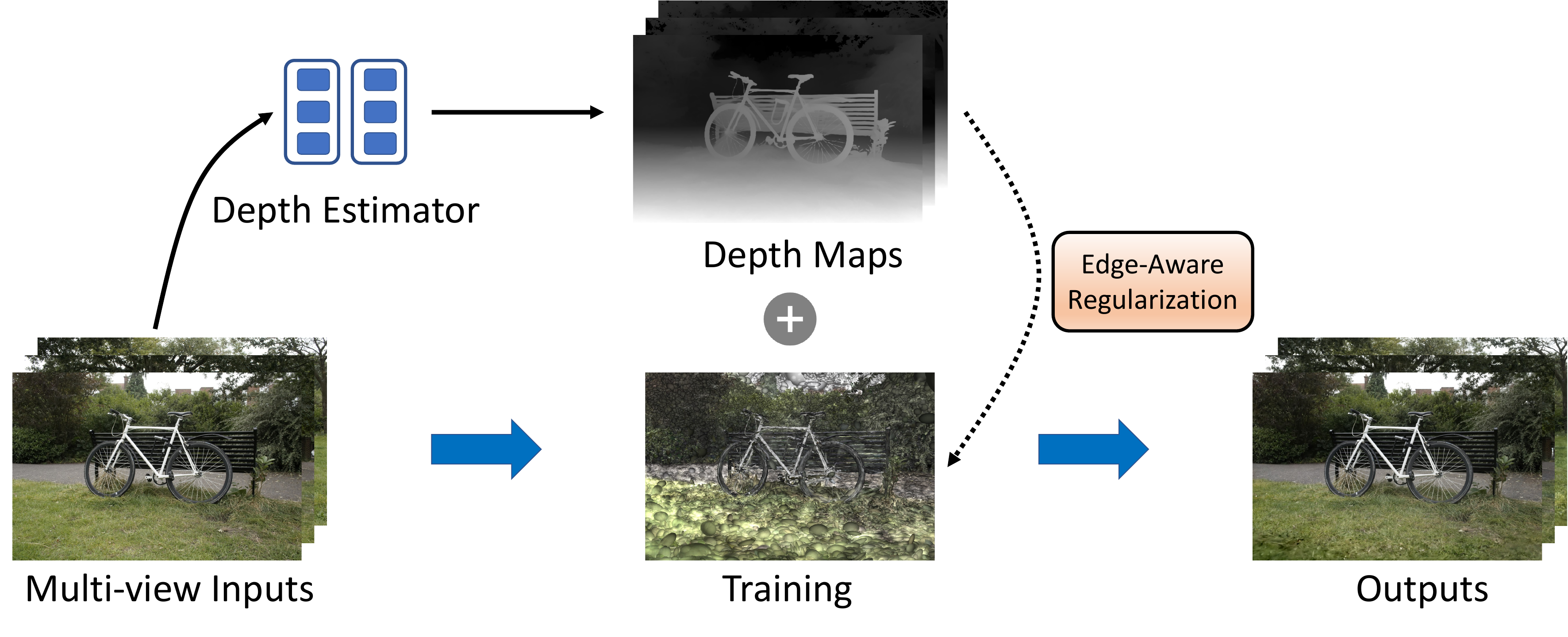}
    \caption{DET-GS: Our method reconstructs 3D scenes from multi-view RGB inputs using a representation of 3D Gaussian primitives. To enhance geometric coherence, we introduce hierarchical depth supervision leveraging depth maps predicted by a pretrained estimator. We further propose an edge-aware regularization strategy based on edge masks extracted via Canny edge detection, preserving essential geometric boundaries. Additionally, we develop an RGB-guided total variation loss to refine appearance details effectively. These novel components collectively enforce geometric consistency, yielding photorealistic renderings that faithfully retain structural detail.}
    \label{fig:intro}
\end{figure*}

To address these challenges, we propose \textbf{DET-GS}, a depth and edge-aware regularization framework for 3D Gaussian Splatting that fundamentally refocuses geometric supervision and structural preservation. DET-GS introduces three key innovations designed to enhance both geometric accuracy and visual fidelity under sparse-view conditions. First, our method introduces a hierarchical geometric depth supervision strategy that integrates an error-tolerant mechanism, delivering robust geometric guidance while effectively mitigating the impact of depth estimation noise and addressing the inherent limitations of conventional non-local regularization schemes. To preserve important geometric structures, we design an edge-aware depth smoothing approach based on Canny edge detection~\cite{canny1986computational}, which selectively regularizes non-boundary regions while maintaining sharp object contours. Finally, we develop an RGB-guided edge-preserving Total Variation loss, a structure-aware image-space regularization technique that selectively enforces smoothness in homogeneous areas while rigorously protecting high-frequency details and texture (see Fig.~\ref{fig:intro}).

Through rigorous experimentation, DET-GS demonstrates substantial improvements over existing state-of-the-art (SOTA) 3DGS-based methods in both geometric accuracy and visual fidelity. Ablation studies further validate the effectiveness of each proposed component, highlighting the importance of hierarchical geometric depth supervision, edge-aware regularization, and structure-preserving image-space smoothing. These results collectively establish DET-GS as a robust and principled solution for high-quality 3D scene reconstruction under sparse-view constraints.

\subsection*{Contributions}
In summary, the contributions of this work are as follows:

\begin{enumerate}
    \item We propose a hierarchical geometric depth supervision strategy that captures geometric structures across multi-level geometric details and introduces an error-tolerant mechanism to improve robustness against noisy monocular depth priors.

    \item We design an edge-aware depth regularization mechanism guided by semantic masks derived from Canny edge detection, enabling selective smoothing that preserves critical scene boundaries and enhances geometric fidelity.

    \item We develop an RGB-guided edge-preserving Total Variation loss that selectively smooths homogeneous regions while rigorously protecting high-frequency textures, significantly improving the perceptual quality of novel view synthesis.
\end{enumerate}
\section{Related Work}
\noindent \textbf{MLP-based Radiance Fields}\ \ \ \
Early neural field methods primarily relied on multi-layer perceptrons (MLPs) to approximate both the geometric structure and visual features of three-dimensional scenes. These models process spatial coordinates and viewing angles to derive scene-specific properties. Typical outputs include the signed distance function (SDF) of the surface geometry~\cite{park2019deepsdf,wang2021neus,wang2023neus2}, or density and color values at particular spatial points~\cite{mildenhall2021nerf,barron2021mip,barron2022mip}. Neural Radiance Field (NeRF) methods~\cite{mildenhall2021nerf} have significantly advanced novel view synthesis by producing highly realistic imagery and effectively modeling view-dependent visual effects. Utilizing established volume rendering formulas, NeRF methods train a coordinate-centric MLP to simultaneously encode geometry and radiance, mapping these directly from spatial coordinates and viewing directions enhanced by positional encoding. Despite delivering high synthesis quality through volumetric modeling, MLP-based approaches confront notable limitations. The primary constraint is the necessity for extensive sampling along each camera ray, necessitating that the MLP assess several sites. This expensive procedure significantly reduces rendering speed and limits performance, particularly for extensive and complex scenes. To mitigate the computational challenges associated with dense sampling and extensive MLP evaluations, various hybrid feature-grid methods have been introduced. These approaches~\cite{yu2021plenoctrees,liu2020neural,fridovich2022plenoxels,chen2022tensorf,muller2022instant,huang2024efficient} essentially ``cache'' intermediate feature representations to streamline the rendering pipeline. Among these techniques, multi-resolution hash encoding~\cite{muller2022instant} has gained popularity, offering a versatile foundation that accelerates rendering by capturing detailed scene information across multiple scales and facilitating efficient level-of-detail (LOD) renderings~\cite{xu2023vr, barron2023zip}. Although these solutions significantly improve rendering speed and visual quality, they still fail to eliminate the fundamental computational burden posed by extensive ray casting operations. Consequently, there remains a substantial performance gap preventing true real-time rendering.

\vspace{1em}

\noindent \textbf{Point-based Radiance Fields}\ \ \ \
Point-based radiance field rendering benefits significantly from using explicit point clouds as proxies for scene representation. These point sets can be quickly and accurately acquired using LiDAR sensors~\cite{liao2022kitti} or through reconstruction techniques such as Structure-from-Motion (SfM) and Multi-View Stereo (MVS)~\cite{schonberger2016pixelwise}. Recent advancements incorporate neural-based descriptors~\cite{ruckert2022adop} or specially designed attributes~\cite{kopanas2021point}, enabling high-quality visual synthesis via differentiable point rendering pipelines~\cite{wiles2020synsin,arvanitis2021broad}. However, discrete rasterization can introduce visual imperfections, notably aliasing and overdrawing, especially when multiple points overlap in a single pixel. Recently, 3D Gaussian Splatting (3DGS)~\cite{kerbl20233d} has significantly enhanced novel view synthesis, achieving real-time rendering at high-definition resolutions. Unlike continuous density-based fields and ray-based methods, 3DGS leverages a collection of anisotropic 3D Gaussians. Rendering occurs through rasterization, projecting these Gaussians onto a 2D image plane. Pixels are then determined by sorting the projected Gaussians based on depth and blending them using alpha composition. Concurrent studies have rapidly expanded the capabilities of the 3DGS framework, focusing on improvements in visual fidelity~\cite{yang2024spec,huang2025structgs,huang2025gaussianfocus,guo2024motion,zhou2025gedr,li2025frpgs,yu2024get3dgs}, avatar reconstruction~\cite{zheng2024gpsgaussian,qian20243dgs}, anti-aliasing techniques~\cite{yu2024mip,yan2024multi,liu2025mesh}, and structured grid representations~\cite{lu2024scaffold}. Nevertheless, these enhancements primarily emphasize visual appearance improvements without explicitly tackling underlying geometric precision.

\vspace{1em}

\noindent \textbf{Depth Supervision in Radiance Fields}\ \ \ \
Depth has previously functioned as a crucial indicator in several 3D computer vision applications~\cite{wang2022uncertainty,wang2024robust,wang2024contrastive}, and has lately been prominent in the supervision of sparse-view radiance fields. Current depth-supervised methodologies may be generally classified into two categories. One category~\cite{deng2022depth,roessle2022dense} obtains accurate but sparse depth information from reliable point cloud sources, whereas another category~\cite{song2023darf,wang2023sparsenerf,yu2022monosdf} generates dense depth cues from dependable monocular depth estimators~\cite{ranftl2021vision,roessle2022dense,yang2024depth1,yang2024depth2}. Monocular depth forecasts provide enhanced reliability and density in instances when point clouds are sparse or absent. Monocular depth assessments are intrinsically plagued by scale ambiguity and possible inaccuracy. To tackle these challenges, previous studies and new sparse-view 3D Gaussian Splatting (3DGS) methods employ scale-invariant loss functions~\cite{song2023darf,xiong2024sparsegs,zhu2024fsgs,li2024dngaussian,10678533,yu2024lm,shen2025dof}, such as depth ranking loss~\cite{wang2023sparsenerf}. Although they are popular, these strategies possess limits within our context. The highly adaptable character of Gaussian representations primarily heightens susceptibility to erroneous depth signals, necessitating more regularization. Moreover, these losses generally impose alignment on a non-local scale, disregarding local depth discrepancies. This neglect can result in noisy Gaussian distributions, particularly in regions with intricate textures. Furthermore, existing regularization methods~\cite{li2024dngaussian,10678533,yu2024lm,shen2025dof} apply uniform smoothing over the entire scene, ignoring semantic and geometric boundaries. This often leads to blurred edges and a loss of fine structural details. To address the limitations discussed above, we propose a hierarchical geometric regularization framework that effectively mitigates noise sensitivity inherent in monocular depth estimation and avoids the shortcomings associated with non-local smoothing. We also propose a depth smoothing strategy guided by Canny edge detection~\cite{canny1986computational}, enabling targeted regularization within homogeneous regions and simultaneously preserving sharp boundaries crucial for maintaining geometric accuracy. Moreover, we devise an RGB-driven, structure-sensitive Total Variation (TV) loss in image space, selectively imposing smoothness only on uniform regions while rigorously safeguarding intricate textures and high-frequency details.
\section{Methodology}

\begin{figure*}[!t]
    \centering
    \includegraphics[width=1\textwidth]{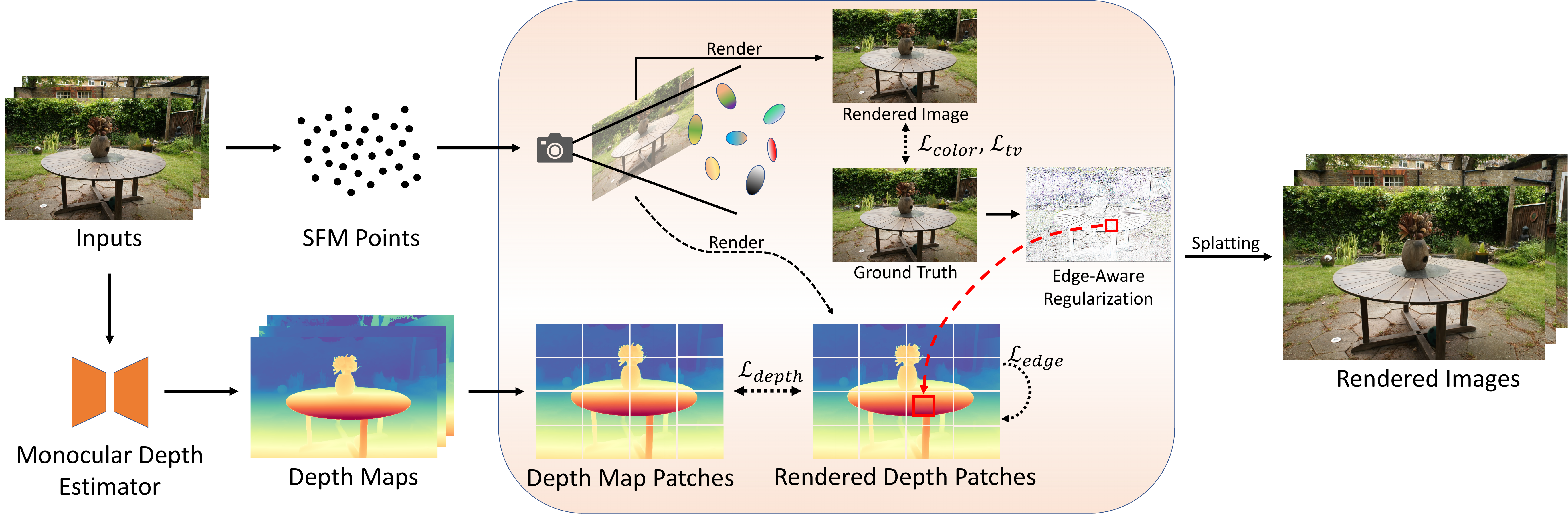}
    \caption{\textbf{Overview of DET-GS:} Given a set of RGB images, a pre-trained monocular depth estimator provides monocular depth priors to guide the optimization. The scene is represented by a set of 3D Gaussian primitives, which can render both color images and depth maps. During training, we decompose both the predicted and pseudo-ground-truth depth maps into non-overlapping patches and apply \textbf{Hierarchical Geometric Depth Supervision}, enabling dual-scale geometric consistency through patch-wise normalization. To preserve geometric boundaries, we incorporate an \textbf{Edge-Aware Depth Regularization} mechanism based on Canny edge detection from ground-truth RGB images, enforcing structure-aware smoothing selectively in non-edge regions. Additionally, we propose a \textbf{RGB-Guided Edge-Preserving Total Variation Regularization}, which adaptively penalizes image-space gradients based on RGB-derived structural information. These loss components ($\mathcal{L}_\text{depth}$, $\mathcal{L}_\text{edge}$, $\mathcal{L}_\text{tv}$, $\mathcal{L}_\text{color}$) collectively guide the optimization of Gaussian parameters for high-fidelity 3D reconstructions.}
    \label{fig:model_structure}
\end{figure*}

\subsection{Preliminary}
3D Gaussian splatting~\cite{kerbl20233d} leverages multiple 3D Gaussian elements to encode spatial information. Specifically, this technique generates the color value $\mathcal{C}$ at each pixel using Gaussian primitives denoted as $\theta$, camera orientation $P$, and intrinsic camera parameters including the center position $o$. Each Gaussian primitive is characterized by three distinct parameters: a central position $\mu \in \mathbb{R}^3$, a scaling vector $s \in \mathbb{R}^3$, and a rotational quaternion $q \in \mathbb{R}^4$. Formally, the $n$-th primitive's Gaussian basis function $\mathcal{G}_n$ is mathematically given by:
\begin{equation}
\mathcal{G}_n(x) = e^{-\frac{1}{2}(x-\mu_n)^T\Sigma_n^{-1}(x-\mu_n)},
\end{equation}
where the covariance $\Sigma_n$ is determined using scale $s_n$ and rotation $q_n$.
Each Gaussian primitive further contains an opacity scalar $\alpha \in \mathbb{R}$ and a color descriptor feature $f \in \mathbb{R}^K$. Consequently, the full parameterization for Gaussian primitive $i$ is denoted as $\theta_n = {\mu_n, s_n, q_n, \alpha_n, f_n}$.
Rendering with 3D Gaussian splatting relies on point-based accumulation. Specifically, pixel color $\mathcal{C}(x_i)$ results from compositing the contributions of $N$ overlapping Gaussians:
\begin{equation}
\mathcal{C}(x_i) = \sum_{n \in N}{c_n\widetilde{\alpha}_n\prod_{j=1}^{n-1}(1-\widetilde{\alpha}_j)}, \text{where}~\widetilde{\alpha}_n = \alpha_n \mathcal{G}_n^{2D}(x_i),
\end{equation}
with $c_n$ representing the decoded color from feature $f_n$. Distinct from conventional radiance field~\cite{mildenhall2021nerf}, which performs ray-wise sampling, Gaussian primitives for each pixel are selected using an optimized rasterization method, guided by pixel coordinate $x_i$, camera intrinsics, pose $P$, and certain predefined criteria. The rendered opacity $\widetilde{\alpha}_n$ of each Gaussian primitive is computed from its inherent opacity $\alpha_n$ and its 2D projection $\mathcal{G}^{2D}_n$ on the image plane. 

Depth value $\mathcal{D}(x_i)$ at each pixel location is calculated analogously, defined by the distance from the primitive center $\mu_n$ to the camera center $o$:
\begin{equation}
\setlength{\abovedisplayskip}{2pt}
\setlength{\belowdisplayskip}{2pt}
\mathcal{D}(x_i) = \sum_{n \in N}{||\mu_n - o||_2} \times \widetilde{\alpha}_n \prod_{j=1}^{n-1}(1-\widetilde{\alpha}_j).
\end{equation}
Parameter optimization in 3D Gaussian splatting is performed through gradient-based methods guided by color information. During the training process, the system duplicates the primitives exhibiting the highest activation to better capture complex visual features, simultaneously pruning unnecessary elements. This study adopts these optimization strategies directly for color guidance.
Traditionally, methods initialize Gaussians with point clouds from COLMAP~\cite{schonberger2016structure,schonberger2016pixelwise} or similar structure-from-motion approaches.
\subsection{Hierarchical Geometric Depth Supervision}
Existing 3DGS-based methods~\cite{10678533,yu2024lm,shen2025dof} typically enforce depth consistency at broad spatial scales, overlooking fine-grained geometric variations within local regions. This oversight leads to suboptimal Gaussian primitive arrangements, especially in areas containing complex surface details where precise depth alignment is crucial. Such non-local-only supervision fails to address localized depth inconsistencies, resulting in scattered Gaussian distributions that compromise reconstruction quality in texture-rich regions.

We leverage Depth Anything V2~\cite{yang2024depth2} as a prior to estimate depth predictions $\widetilde{\mathcal{D}}$ for training images. It is a state-of-the-art (SOTA) pre-trained monocular depth estimation model. This provides pseudo-ground-truth depth information that guides the optimization of Gaussian primitive positions during training. During the training process, we compute the predicted depth $\mathcal{D}$ from our 3D Gaussian representation by rendering depth values along rays cast from the camera center $o$ through each pixel $x_i$. To effectively utilize depth information across different spatial scales, we decompose both rendered images and corresponding depth maps into non-overlapping patches, as shown in Fig.~\ref{fig:model_structure}. This approach enables fine-grained depth supervision at multiple hierarchical levels, allowing the model to capture both local geometric details and global structural coherence. For spatial reshaping of Gaussian fields, we compute a patch-wise depth that emphasizes the contribution of nearest Gaussians by applying enhanced opacity values $\omega$ to all primitives:
\begin{equation}
\begin{split}
\mathcal{D}(x_i) = \sum_{n \in N}{\omega \exp(-\iota(n-1)) \mathcal{G}^{2D}_n(x_i) ||\mu_n - o||_2 }, \\ \text{where}~\iota=-ln(1-\omega),
\end{split}
\end{equation}
where $\mu_i$ denotes the center position of the $n$-th Gaussian primitive. The term $-ln(1-\omega)$ makes the nearest Gaussian receive the highest weight, and farther Gaussians contribute less proportionally.

Our approach establishes a dual-scale geometric supervision paradigm that jointly optimizes hierarchical spatial relations and enhances structural fidelity across scales. For each patch $\mathcal{P}$, we focus on capturing relative depth variations within localized regions to enhance fine-grained geometric structure learning.:
\begin{equation}
\begin{split}
    \mathcal{D}_{patch}(x) = \frac{\mathcal{D}(x) - \mu_{\mathcal{P}}}{\sigma_{\mathcal{P}} + \delta}, \text{where}~\mu_{\mathcal{P}} = \frac{1}{|\mathcal{P}|}\sum_{x \in \mathcal{P}}\mathcal{D}(x) \\~\text{and}~\sigma_{\mathcal{P}} = \sqrt{\frac{1}{|\mathcal{P}|}\sum_{x \in \mathcal{P}}(\mathcal{D}(x) - \mu_{\mathcal{P}})^2}~,
\end{split}
\end{equation}
where $x \in \mathcal{P}$ and $\delta$ ensures numerical stability. The proposed fine-scale supervision strategy accentuates subtle depth variations that are typically diminished in coarse-grained processing, ensuring accurate reconstruction of geometric structures. To maintain overall scene coherence, we apply an image-based normalization using image-wide statistics:
\begin{equation}
\begin{aligned}
\mathcal{D}_{image}(x) &= \frac{\mathcal{D}(x) - \mu_{\mathcal{P}}}{\sigma_{\mathcal{I}}}, \\
\text{where}~\sigma_{\mathcal{I}} &= \sqrt{\frac{1}{|\mathcal{I}|} \sum_{x \in \mathcal{I}} (\mathcal{D}(x) - \mu_{\mathcal{I}})^2}~\text{and}~\mu_{\mathcal{I}} = \frac{1}{|\mathcal{I}|} \sum_{x \in \mathcal{I}} \mathcal{D}(x).
\end{aligned}
\end{equation}
Here, while maintaining patch-wise mean subtraction, we utilize the global standard deviation from the entire image $\mathcal{I}$. This preserves the global depth distribution while allowing local adaptivity. The depth supervision directly targets the mean positions $\mu$ of 3D Gaussian primitives. By computing the loss between rendered depth maps and Depth Anything V2 predictions $\widetilde{\mathcal{D}}$ at both scales, we create optimization gradients that drive Gaussian centers toward geometrically consistent positions.

The final depth supervision loss combines both hierarchical scales through similarity losses computed on target areas:
\begin{equation}
    \mathcal{L}_{depth} = \gamma \|\mathcal{D}_{patch} - \widetilde{\mathcal{D}}_{patch}\|_2^2 + \eta \|\mathcal{D}_{image} - \widetilde{\mathcal{D}}_{image}\|_2^2,
\end{equation}
where $\widetilde{D}_{patch}$ and $\widetilde{D}_{image}$ represent the correspondingly normalized depth predictions from monocular depth estimator~\cite{yang2024depth2}, and $\gamma$, $\eta$ are weighting factors balancing local and global supervision contributions. This hierarchical loss formulation enables the model to learn accurate geometry at multiple scales simultaneously, resulting in improved 3D scene reconstruction quality.
\subsection{Edge-Aware Depth Regularization}
Accurately reconstructing fine geometric structures requires preserving sharp depth discontinuities at object boundaries, where conventional uniform regularization often leads to undesirable blurring and structural degradation. To overcome this fundamental limitation, we propose a novel edge-aware depth regularization strategy that integrates structural priors from the RGB image, enabling geometry-adaptive smoothness control. Specifically, we leverage the inherent correlation between image edges and depth discontinuities. A structural edge map $\mathcal{E}$ was extracted using the Canny edge detector~\cite{canny1986computational} from the ground-truth RGB image $\mathcal{I}$:
\begin{equation}
\mathcal{E} = \text{Canny}(\mathcal{I}),
\end{equation}
where the lower and upper thresholds for Canny detection are empirically set to 20 and 200, ensuring a robust yet sensitive edge response that captures critical structural boundaries. To focus smoothing on homogeneous regions, we construct a non-edge mask by inverting $\mathcal{E}$. We define an edge-aware binary mask $m(x_i)$ as:
\begin{equation}
m(x_i) = 
\begin{cases}
1, & \text{if } \mathcal{E}(x_i) = 0, \\
0, & \text{otherwise}.
\end{cases}
\end{equation}
This binary mask ensures that regularization is applied only to homogeneous regions, while structural boundaries are explicitly preserved. Unlike traditional depth smoothness terms that uniformly enforce local continuity, our method selectively constrains smoothness only within non-edge regions. Let $\mathcal{D}(x)$ denote the rendered depth at pixel $x$. For each pixel $x_i$, we define its local neighborhood $\mathcal{N}(x_i)$ and compute the masked local mean depth as:
\begin{equation}
\overline{\mathcal{D}}(x_i) = \frac{ \sum_{x_j \in \mathcal{N}(x_i)} \mathcal{D}(x_j) \cdot m(x_j) }{ \sum_{x_j \in \mathcal{N}(x_i)} m(x_j) + \epsilon },
\end{equation}
where $\overline{\mathcal{D}}(x_i)$ represents the masked local mean depth within a neighborhood centered at $x_i$, and $\epsilon=10^{-8}$ is a small constant to ensure numerical stability. We adopt a cross-shaped $3\times3$ convolutional kernel (consisting of the center and four-connected neighbors) to balance fine-grained detail preservation and smoothing efficiency while reducing the risk of over-smoothing near object boundaries. The proposed edge-aware depth regularization loss is formulated as:
\begin{equation}
\mathcal{L}_{edge} = \frac{1}{P} \sum_{x_i} m(x_i) \cdot \left| \mathcal{D}(x_i) - \overline{\mathcal{D}}(x_i) \right|^2,
\end{equation}
where $P$ denotes the number of valid pixels considered for regularization. This strategy establishes an explicit synergy between RGB structures and 3D depth geometry. By incorporating structural priors into the depth regularization process, we enable a content-aware smoothing mechanism that dynamically adapts to the underlying scene geometry. Unlike traditional uniform smoothing, our edge-aware approach employs a structure-guided, masked, cross-shaped local mean constraint that preserves boundary sharpness and enhances geometric fidelity under sparse-view constraints.
\subsection{RGB-Guided Edge-Preserving Total Variation Regularization}
While depth-based supervision focuses on recovering accurate 3D structures, ensuring smoothness and preserving visual details in the rendered images remains crucial for achieving high-fidelity reconstructions. Conventional Total Variation (TV) regularization promotes spatial smoothness by penalizing large intensity differences between adjacent pixels. However, standard TV loss tends to over-smooth edges and fine structures, resulting in the loss of critical details in the reconstructed images. To address this limitation, we propose an RGB-guided edge-preserving Total Variation regularization that selectively enforces smoothness constraints based on the local gradient information from the ground-truth RGB image.
Let $\mathcal{I}_{\text{pred}}$ denote the rendered image and $\mathcal{I}_{\text{gt}}$ the ground-truth image, both defined over pixel grid $\Omega$. $\Omega$ denotes the set of all pixel coordinates in the image domain, formally defined as:
\begin{equation}
\Omega = \left\{ (a, b) \mid 1 \leq a \leq H,\ 1 \leq b \leq W \right\},
\end{equation}
where $H$ and $W$ are the height and width of the image, respectively. For each pixel $x_i \in \Omega$, we define the horizontal and vertical gradients for the predicted and ground-truth images as:
\begin{equation}
\begin{aligned}
\nabla_h \mathcal{I}_{\text{pred}}(x_i) &= \mathcal{I}_{\text{pred}}(x_i + 1_h) - \mathcal{I}_{\text{pred}}(x_i), \\
\nabla_h \mathcal{I}_{\text{gt}}(x_i) &= \mathcal{I}_{\text{gt}}(x_i + 1_h) - \mathcal{I}_{\text{gt}}(x_i), \\
\nabla_v \mathcal{I}_{\text{pred}}(x_i) &= \mathcal{I}_{\text{pred}}(x_i + 1_v) - \mathcal{I}_{\text{pred}}(x_i), \\
\nabla_v \mathcal{I}_{\text{gt}}(x_i) &= \mathcal{I}_{\text{gt}}(x_i + 1_v) - \mathcal{I}_{\text{gt}}(x_i),
\end{aligned}
\end{equation}
where $1_h$ and $1_v$ denote one-pixel shifts in the horizontal and vertical directions, respectively. To avoid penalizing genuine edges and sharp transitions, we introduce an edge-aware modulation function based on the RGB gradients. Specifically, for each direction, we construct binary masks $M_h(x_i)$ and $M_v(x_i)$ as:
\begin{equation}
\begin{aligned}
M_h(x_i) &= \mathbb{I}\left( \left| \nabla_h \mathcal{I}_{\text{gt}}(x_i) \right| < \tau_{\text{edge}} \right), \\
M_v(x_i) &= \mathbb{I}\left( \left| \nabla_v \mathcal{I}_{\text{gt}}(x_i) \right| < \tau_{\text{edge}} \right),
\end{aligned}
\end{equation}
where $\tau_{\text{edge}}$ is a threshold controlling edge sensitivity, and $\mathbb{I}(\cdot)$ denotes the indicator function, returning 1 if the condition holds and 0 otherwise. The edge-preserving Total Variation loss is then defined as:
\begin{equation}
\begin{aligned}
\mathcal{L}_{\text{tv}} = \frac{1}{|\Omega|} \sum_{x \in \Omega} \Big(
& M_h(x) \cdot \max\left( \left| \nabla_h \mathcal{I}_{\text{pred}}(x) \right| - \tau_{\text{smooth}},\, 0 \right) \\
& + M_v(x) \cdot \max\left( \left| \nabla_v \mathcal{I}_{\text{pred}}(x) \right| - \tau_{\text{smooth}},\, 0 \right) 
\Big),
\end{aligned}
\end{equation}
where $\tau_{\text{smooth}}$ is a small margin to tolerate minor gradient fluctuations. We set $\tau_{\text{edge}}=10^{-2}$ to filter out strong edges and $\tau_{\text{smooth}}=10^{-4}$ to suppress small gradients. The final TV regularization term is scaled and integrated into the overall optimization objective. This formulation promotes smoothness in homogeneous regions while preserving edges and fine details aligned with the ground-truth structure. By adapting the regularization strength to local image gradients, our method achieves a better balance between noise suppression and detail preservation. In contrast to conventional TV loss that uniformly penalizes intensity differences, our RGB-guided strategy leverages structural cues to selectively smooth flat areas without degrading important textures and boundaries, resulting in more photorealistic rendering.
\subsection{Training Losses}
The overall training objective of our method consists of four components: a color reconstruction loss $\mathcal{L}_{\text{color}}$, a hierarchical geometric depth supervision loss $\mathcal{L}_{\text{depth}}$, an edge-aware depth regularization loss $\mathcal{L}_{\text{edge}}$, and an RGB-guided edge-preserving total variation loss $\mathcal{L}_{\text{tv}}$. Following 3D Gaussian Splatting~\cite{kerbl20233d}, the color reconstruction loss combines an $L_1$ reconstruction loss and a D-SSIM term between the rendered image $\mathcal{I}_{\text{pred}}$ and the ground-truth image $\mathcal{I}_{\text{gt}}$:
\begin{equation}
    \mathcal{L}_{color} = \mathcal{L}_1({\mathcal{I}_\text{pred}}, \mathcal{I}_\text{gt}) + \lambda \mathcal{L}_{\mathrm{D-SSIM}}({\mathcal{I}_\text{pred}}, \mathcal{I}_\text{gt}) .
\end{equation}
The final training loss function is defined as:
\begin{equation}
\mathcal{L} = \mathcal{L}_{\text{color}} + \mathcal{L}_{\text{depth}} + \beta \mathcal{L}_{\text{edge}} + \phi \mathcal{L}_{\text{tv}}.
\end{equation}

\begin{table*}[!htbp]
\centering
\caption{\textbf{Quantitative Comparison Results on the Mip-NeRF 360~\cite{barron2022mip}, Tanks\&Temples~\cite{knapitsch2017tanks} and DeepBlending~\cite{hedman2018deep} Datasets.} Our method outperforms baselines and SOTA methods~\cite{kerbl20233d,li2024dngaussian,yu2024mip,yang2024spec}. Some competing metrics are sourced from the respective papers.}
\label{tab:mip_1}
\begin{adjustbox}{width=\textwidth,center}
\begin{tabular}{lccc|ccc|ccc}
\hline
& \multicolumn{3}{c|}{Mip-NeRF 360} & \multicolumn{3}{c|}{Tanks\&Temples} & \multicolumn{3}{c}{Deep Blending} \\
& PSNR $\uparrow$ & SSIM $\uparrow$ & LPIPS $\downarrow$ & PSNR $\uparrow$ & SSIM $\uparrow$ & LPIPS $\downarrow$ & PSNR $\uparrow$ & SSIM $\uparrow$ & LPIPS $\downarrow$  \\
\hline
Instant-NGP~\cite{muller2022instant} & 25.59 & 0.699 & 0.331 & 21.72 & 0.723 & 0.330 & 23.62 & 0.797 & 0.423 \\
Plenoxels~\cite{fridovich2022plenoxels} &23.08 & 0.626 & 0.463 & 21.08 & 0.719 & 0.379 & 23.06 & 0.795 & 0.510 \\
Mip-NeRF 360~\cite{barron2022mip} & 27.69 & 0.792 & 0.237 & 22.22 & 0.759 & 0.257 & 29.40 & 0.901 & 0.245 \\
\hline
3DGS~\cite{kerbl20233d} & 27.79 & 0.826 & 0.202 & 23.13 & 0.840 & 0.182 & 29.42 & 0.905 & 0.241 \\
DNGaussian~\cite{li2024dngaussian} & 27.87 & 0.828 & 0.195 & 23.75 & 0.846 & 0.178 & 29.65 & 0.904 & 0.240 \\
Mip-Splatting~\cite{yu2024mip} & 27.79 & 0.827 & 0.203 & 23.86 & 0.853 & 0.175 & 29.71 & 0.904 & 0.242 \\
Scaffold-GS~\cite{lu2024scaffold} & 27.98 & 0.824 & 0.207 & 23.93 & 0.854 & 0.176 & \textbf{30.20} & 0.905 & 0.253\\
Spec-Gaussian~\cite{yang2024spec} & 28.12 & 0.834 & 0.177 & 23.79 & 0.858 & 0.166 & 29.75 & 0.906 & 0.241 \\
\hline
Ours & \textbf{28.29} & \textbf{0.840} & \textbf{0.175} & \textbf{23.98} & \textbf{0.861} & \textbf{0.165} & 30.05 & \textbf{0.908} & \textbf{0.239} \\
\hline
\end{tabular}
\end{adjustbox}
\end{table*}

\begin{table*}[ht]
\centering
\begin{minipage}[ht]{0.49\textwidth}
\centering
\caption{\textbf{Results on NeRF synthetic dataset~\cite{mildenhall2021nerf}. Some competing metrics are sourced from the respective papers.}} 
\label{tab:general-nerf-comparison}
\begin{tabular}{l|ccc}
\hline  & \multicolumn{3}{c}{ NeRF Synthetic } \\
 & PSNR  $\uparrow$  & SSIM  $\uparrow$  & LPIPS  $\downarrow$\\
\hline 
Instant-NGP~\cite{muller2022instant} & 33.18 & 0.963 & 0.045 \\
Mip-NeRF~\cite{barron2021mip} & 33.09 & 0.961 & 0.043 \\
Tri-MipRF~\cite{hu2023tri} & 33.65 & 0.963 & 0.042 \\
\hline
3D-GS~\cite{kerbl20233d} & 33.78 & 0.969 & 0.031 \\
DNGaussian~\cite{li2024dngaussian} & 33.81 & 0.968 & 0.030 \\
Mip-Splatting~\cite{yu2024mip} & 33.88 & 0.970 & 0.032 \\
Scaffold-GS~\cite{lu2024scaffold} & 33.46 & 0.967 & 0.035 \\
Spec-Gaussian~\cite{yang2024spec} & 34.05 & 0.970 & 0.029 \\
\hline
Ours & \textbf{34.13} & \textbf{0.972} & \textbf{0.028} \\
\hline
\end{tabular}
\end{minipage}
\hfill
\begin{minipage}[ht]{0.49\textwidth}
\centering
\caption{\textbf{Results on NSVF synthetic~\cite{liu2020neural} dataset. Some competing metrics are sourced from the respective papers.}} 
\label{tab:general-nsvf-comparison}
\begin{tabular}{l|ccc}
\hline  & \multicolumn{3}{c}{ NSVF Synthetic } \\
 & PSNR  $\uparrow$  & SSIM  $\uparrow$  & LPIPS  $\downarrow$\\
\hline 
TensoRF~\cite{chen2022tensorf} & 36.52 & 0.982 & 0.026 \\
Tri-MipRF~\cite{hu2023tri} & 34.58 & 0.973 & 0.030 \\
NeuRBF~\cite{chen2023neurbf} & 37.80 & 0.986 & 0.019  \\
\hline
3D-GS~\cite{kerbl20233d} & 37.01 & 0.987 & 0.016 \\
DNGaussian~\cite{li2024dngaussian} & 37.25 & 0.985 & 0.015 \\
Mip-Splatting~\cite{yu2024mip} & 37.78 & 0.986 & 0.017 \\
Scaffold-GS~\cite{lu2024scaffold} & 36.41 & 0.983 & 0.019  \\
Spec-Gaussian~\cite{yang2024spec} & 38.30 & 0.987 & 0.013 \\
\hline
Ours & \textbf{38.39} & \textbf{0.988} & \textbf{0.012} \\
\hline
\end{tabular}
\end{minipage}
\end{table*}

\section{Experiments}
\subsection{Dataset and Metrics}
To comprehensively evaluate the effectiveness of our method, we conduct experiments on five publicly available datasets, including three real-world datasets and two synthetic ones. These benchmarks are widely adopted in the field of novel view synthesis and are used to assess the performance of baseline methods such as 3DGS~\cite{kerbl20233d}, Scaffold-GS~\cite{lu2024scaffold}, Mip-Splatting~\cite{yu2024mip}, DNGaussian~\cite{li2024dngaussian}, and Spec-Gaussian~\cite{yang2024spec}, which represent the current state of the art. Our evaluation includes nine scenes from the Mip-NeRF 360 dataset   ~\cite{barron2022mip}, two scenes each from the Tanks \& Temples~\cite{knapitsch2017tanks} and DeepBlending~\cite{hedman2018deep} datasets, as well as eight scenes from both the NeRF Synthetic~\cite{mildenhall2021nerf} and NSVF Synthetic~\cite{liu2020neural} datasets. These scenes were selected to cover a broad spectrum of geometric and photometric complexity, enabling us to assess the model’s robustness under varying conditions. The dataset collection spans both indoor and outdoor environments and includes scenes with diverse levels of detail, providing a rigorous test of rendering adaptability and scalability. Moreover, the benchmark suite consists of both real-world captures and synthetic data.

Performance is evaluated using standard metrics for image-based rendering: PSNR (Peak Signal-to-Noise Ratio), SSIM (Structural Similarity Index)~\cite{wang2004image}, and LPIPS (Learned Perceptual Image Patch Similarity)~\cite{zhang2018unreasonable}. These metrics respectively measure pixel-level fidelity, structural coherence, and perceptual quality between the rendered and ground-truth images.
\subsection{Baselines and Implementation}
We evaluate our method against several state-of-the-art baselines, including 3DGS~\cite{kerbl20233d}, Scaffold-GS~\cite{lu2024scaffold}, Mip-Splatting~\cite{yu2024mip}, DNGaussian~\cite{li2024dngaussian}, and Spec-Gaussian~\cite{yang2024spec}, with the latter showing top-tier performance in recent benchmarks. To further broaden the comparison, we also include results from additional representative methods: Instant-NGP~\cite{muller2022instant}, Plenoxels~\cite{fridovich2022plenoxels}, Mip-NeRF~\cite{barron2021mip}, Mip-NeRF 360~\cite{barron2022mip}, Tri-MipRF~\cite{hu2023tri}, TensoRF~\cite{chen2022tensorf}, and NeuRBF~\cite{chen2023neurbf}. All models are trained for 30k iterations to ensure consistency across evaluations.

Our approach is implemented on top of the official PyTorch-based 3D Gaussian Splatting framework. The proposed Hierarchical Geometric Depth Supervision is invoked every 5 iterations. To mitigate over-constraining the learning process, we introduce an error tolerance mechanism into the depth L2 loss. The loss function parameters are fixed across all experiments: $\gamma = 0.1$, $\eta = 1$, $\beta = 0.1$, and $\phi = 0.8$. For experiments on the Mip-NeRF 360 dataset~\cite{barron2022mip}, input images are downsampled by a factor of 4 for outdoor scenes and by a factor of 2 for indoor ones.

\begin{figure*}[!t]
    \centering
    \includegraphics[width=1\textwidth]{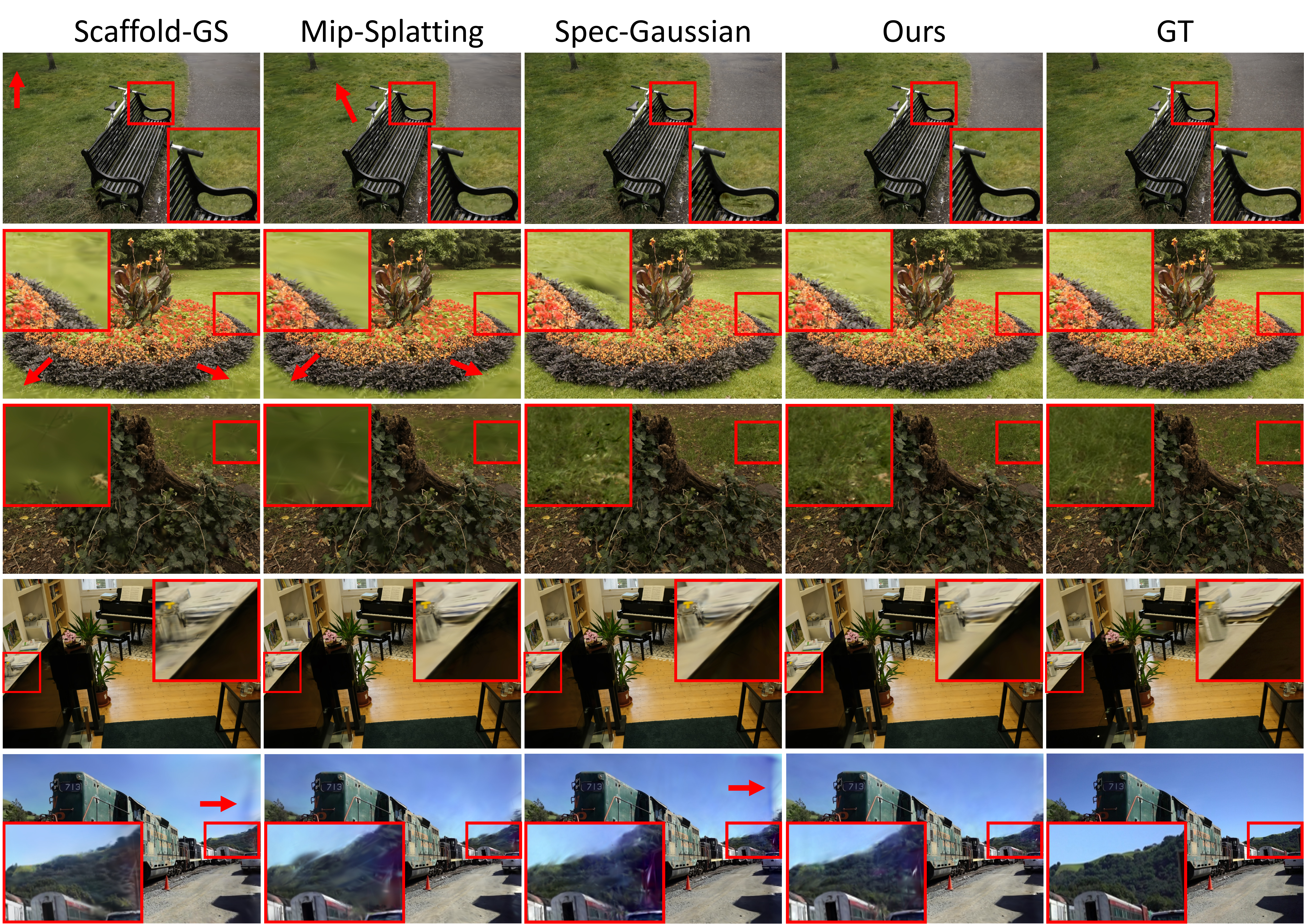}
    \caption{\textbf{Qualitative Comparison Results Across Different Datasets~\cite{barron2022mip,knapitsch2017tanks}.} Zoomed-in regions showcase fine-grained rendering differences. Red arrows mark visual artifacts, such as local blurriness, present in current state-of-the-art~\cite{lu2024scaffold,yu2024mip,yang2024spec} approaches. Compared to these methods, our model achieves more faithful detail preservation and delivers more realistic and high-fidelity renderings.}
    \label{fig:exp_mip}
\end{figure*}

\begin{figure*}[!t]
    \centering
    \includegraphics[width=1\textwidth]{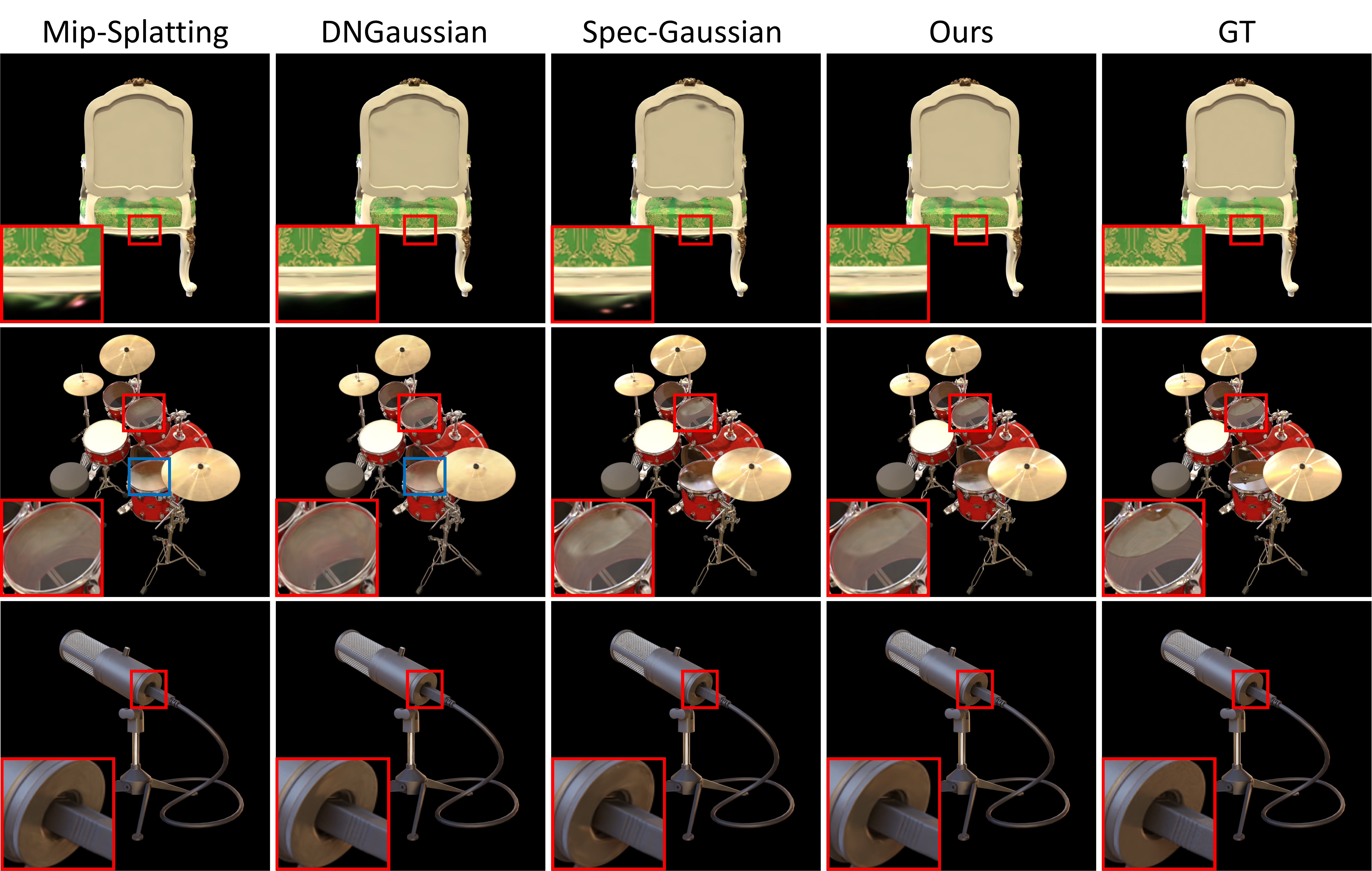}
    \caption{\textbf{Qualitative Comparison Results of NeRF Synthetic Dataset~\cite{mildenhall2021nerf}.} The red boxes highlight detailed regions of the rendered images. With the aid of depth and edge-aware supervision, as well as our proposed TV loss, our approach better preserves intricate visual structures and delivers more accurate and detailed renderings than leading contemporary methods.}
    \label{fig:exp_nerf}
\end{figure*}

\begin{figure*}[ht]
    \centering
    \includegraphics[width=1\textwidth]{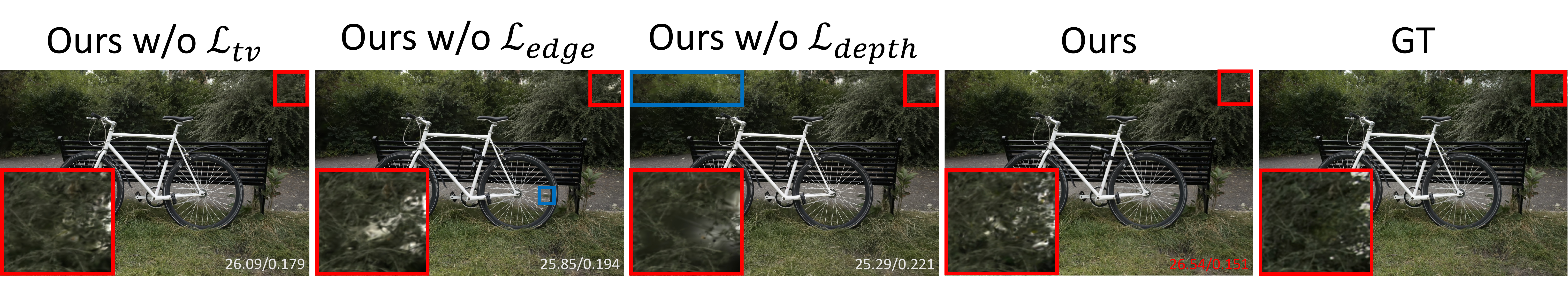}
    \caption{\textbf{Ablation of Our Model Components.} We conduct experiments that remove each of the three core innovations individually to assess their effectiveness. We present the PSNR and LPIPS metrics, with the best scores highlighted in red.}
    \label{fig:ablation}
\end{figure*}

\begin{figure}[t]
    \centering
    \includegraphics[width=0.49\textwidth]{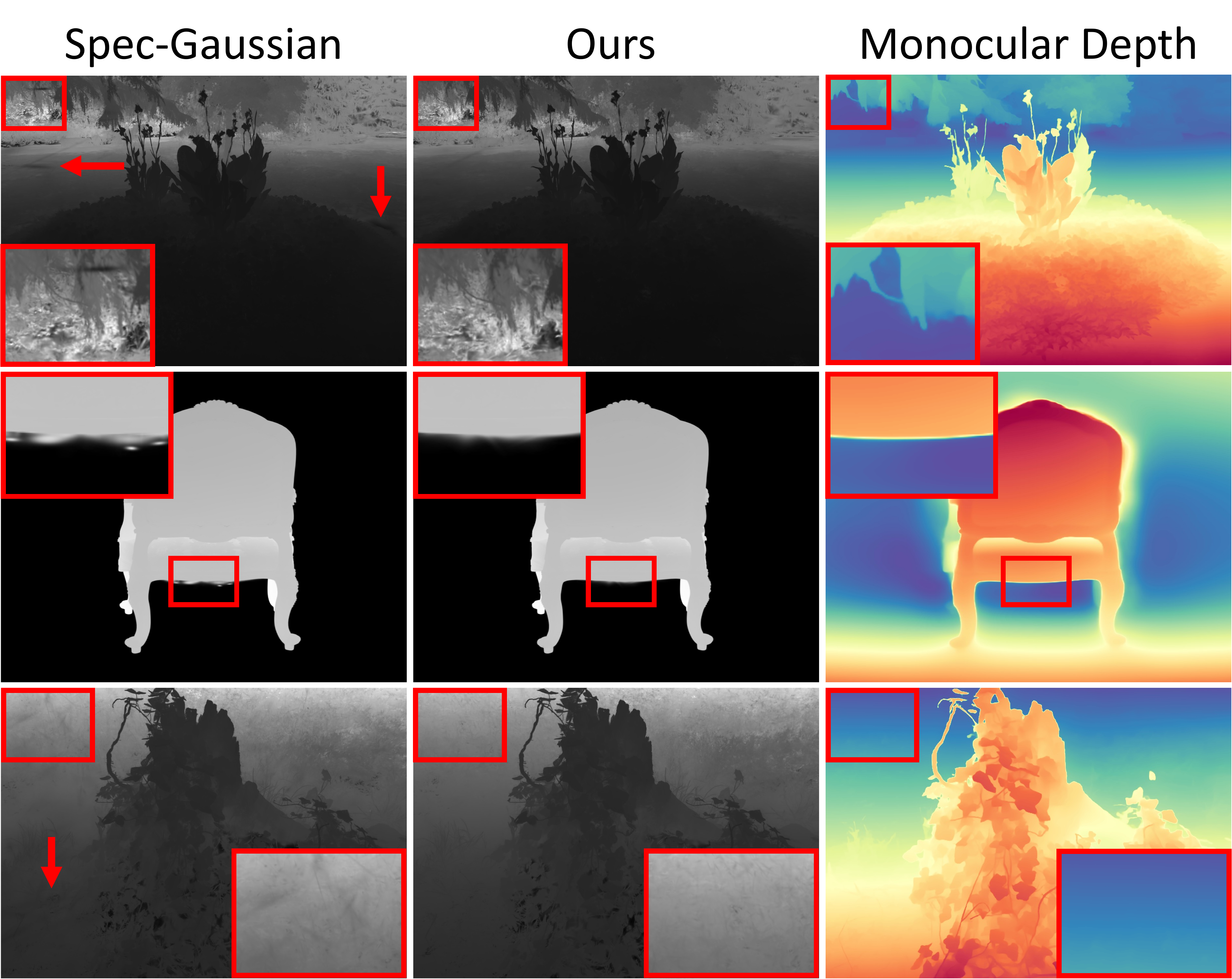}
    \caption{\textbf{Depth Map Comparisons Between Our Method and State-of-the-art models~\cite{yang2024spec}.} While Spec-Gaussian exhibits blurry floaters and less accurate geometric details, our model generates sharper and structurally consistent depth maps. This improvement is attributed to our Hierarchical Geometric Depth Supervision and Edge-Aware Depth Regularization, which guide depth refinement through local-global constraints and edge-preserving smoothing.}
    \label{fig:exp_depth}
\end{figure}

\subsection{Results Comparison}
\paragraph{\textbf{Real-world Unbounded Datasets}}
To evaluate the effectiveness of our method in real-world scenarios, we conduct experiments on the Mip-NeRF 360~\cite{barron2022mip}, Tanks and Temples~\cite{knapitsch2017tanks}, and DeepBlending~\cite{hedman2018deep} datasets. All baseline models are trained using their official default configurations to ensure a fair comparison. As shown in Tab.~\ref{tab:mip_1}, our approach consistently outperforms previous state-of-the-art methods~\cite{kerbl20233d,lu2024scaffold,yu2024mip,li2024dngaussian,yang2024spec} across all datasets. Qualitative results shown in Fig.~\ref{fig:exp_mip} further demonstrate the visual benefits of our method. Compared to existing models, our approach produces sharper details and higher rendering quality results. This performance gain can be attributed to the integration of depth-aware supervision, which refines the spatial positioning of Gaussians in the early training stages. With improved positional accuracy, the Gaussians can be further optimized in subsequent steps to capture scene-specific features better. This results in more accurate rendering with fewer floating artifacts and improved structural consistency, particularly in complex indoor and outdoor scenes. These results highlight that our approach demonstrates a strong ability to preserve fine-grained details and achieve high-fidelity reconstructions across diverse real-world environments.

\paragraph{\textbf{Synthetic Bounded Datasets}}
To validate the effectiveness of our method under controlled conditions, we evaluate it on two widely used synthetic datasets: NeRF Synthetic~\cite{mildenhall2021nerf} and NSVF~\cite{liu2020neural}. The datasets exhibit clean geometry, consistent lighting, and dense ground-truth views, making them appropriate for evaluating fine-grained reconstruction performance. Our comparisons were made with the most relevant state-of-the-art methods, including point-based renderers such as 3DGS~\cite{kerbl20233d}, Scaffold-GS~\cite{lu2024scaffold}, Mip-Splatting~\cite{yu2024mip}, DNGaussian~\cite{li2024dngaussian} and Spec-Gaussian~\cite{yang2024spec}, as well as several NeRF-based approaches such as TensoRF~\cite{chen2022tensorf}, NeuRBF~\cite{chen2023neurbf}, and Tri-MipRF~\cite{hu2023tri}. As shown in Tab.~\ref{tab:general-nerf-comparison} and Tab.~\ref{tab:general-nsvf-comparison}, our method consistently outperforms all baselines in terms of PSNR, SSIM, and LPIPS. This indicates that our depth and edge aware regularization not only enhances visual fidelity but also improves generalization in synthetic environments. Visual results in Fig.~\ref{fig:exp_nerf} demonstrate clearer object contours and sharper edges, particularly in complex scenes. These improvements are attributed to better Gaussian alignment with geometry, driven by hierarchical depth supervision and edge-preserving regularization. In addition to per-view image quality, we also observe improved multi-view consistency. Since our approach enforces coherent structure during optimization, it leads to fewer view-dependent artifacts and more stable reconstructions across adjacent viewpoints. This is especially evident in specular and thin-structure regions, where baseline methods often struggle with geometry misalignment or color bleeding. Together, these results highlight the strength of our approach in maintaining both photometric accuracy and geometric consistency across viewpoints, validating its robustness in synthetic bounded scenarios.

\paragraph{\textbf{Depth Map Comparison}}
Fig.~\ref{fig:exp_depth} presents a comparison of rendered depth maps between our method and existing state-of-the-art approaches. Notably, the depth maps produced by Spec-Gaussian~\cite{yang2024spec} contain visible floaters and exhibit reduced accuracy in object boundaries and fine geometry. In contrast, our method delivers clearer and more structurally faithful depth maps. This is largely due to the proposed Hierarchical Geometric Depth Supervision, which enforces both local and global consistency during training. Additionally, the Edge-Aware Depth Regularization, guided by Canny edge detection, restricts smoothing to homogeneous regions while preserving sharp transitions at semantic boundaries. Unlike conventional depth regularization that uniformly enforces continuity, our approach selectively encourages smoothness only where it is structurally appropriate.

\subsection{Ablation Study}

\begin{table*}[th]
\centering
\caption{\textbf{Ablation Study of Components in Our Model.} We present quantitative results for two real-world datasets~\cite{barron2022mip,knapitsch2017tanks} and one synthetic dataset~\cite{mildenhall2021nerf} in our ablation study.}
\label{tab:ablation}
\renewcommand{\arraystretch}{1} 
\begin{tabular}{lccc|ccc|ccc}
\hline
& \multicolumn{3}{c|}{Mip-NeRF 360} & \multicolumn{3}{c|}{Tanks\&Temples} & \multicolumn{3}{c}{NeRF Synthetic} \\
& PSNR $\uparrow$ & SSIM $\uparrow$ & LPIPS $\downarrow$ & PSNR $\uparrow$ & SSIM $\uparrow$ & LPIPS $\downarrow$ & PSNR $\uparrow$ & SSIM $\uparrow$ & LPIPS $\downarrow$  \\
\hline
None & 27.69 & 0.823 & 0.201 & 23.07 & 0.837 & 0.185 & 33.79 & 0.968 & 0.033 \\
w/ $\mathcal{L}_{\text{depth}}$ & 28.20 & 0.834 & 0.183 & 23.88 & 0.853 & 0.173 & 34.07 & 0.971 & 0.029 \\
w/ $\mathcal{L}_{\text{edge}}$ & 28.09 & 0.828 & 0.188 & 23.69 & 0.846 & 0.178 & 33.99 & 0.970 & 0.031 \\
w/ $\mathcal{L}_{\text{tv}}$ & 27.89 & 0.825 & 0.195 & 23.57 & 0.841 & 0.181 & 33.90 & 0.970 & 0.032 \\
Ours & \textbf{28.29} & \textbf{0.840} & \textbf{0.175} & \textbf{23.98} & \textbf{0.861} & \textbf{0.165} & \textbf{34.13} & \textbf{0.972} & \textbf{0.028} \\
\hline
\end{tabular}
\end{table*}

To better understand the contributions of each component in our framework, we conduct an ablation study across three datasets: Mip-NeRF 360~\cite{barron2022mip}, Tanks \& Temples~\cite{knapitsch2017tanks}, and NeRF Synthetic~\cite{mildenhall2021nerf}. We evaluate four variants of our model: without Hierarchical Geometric Depth Supervision ($\mathcal{L}_{\text{depth}}$), without Edge-Aware Depth Regularization ($\mathcal{L}_{\text{edge}}$), without RGB-Guided Edge-Preserving TV Regularization ($\mathcal{L}_{\text{tv}}$), and the full model with all components enabled. The results are summarized in Tab.~\ref{tab:ablation}.

\paragraph{\textbf{Hierarchical Geometric Depth Supervision}}
Among all components, removing $\mathcal{L}_{\text{depth}}$ leads to the most significant performance drop across all datasets. This validates the critical role of hierarchical depth cues in guiding the optimization of Gaussian positions. By leveraging both local patch-wise and global image-level depth normalization, this supervision strategy effectively improves geometric alignment, enabling the model to better capture scene structure. As shown in Fig.~\ref{fig:ablation} (bicycle scene, Mip-NeRF 360), the absence of depth supervision causes substantial degradation: floating artifacts become prominent, and scene geometry appears misaligned. The PSNR and LPIPS scores also drop markedly, indicating a loss in both numerical and perceptual fidelity.

\paragraph{\textbf{Edge-Aware Depth Regularization}}
When $\mathcal{L}_{\text{edge}}$ is removed, the rendering quality moderately degrades. Without the guidance of image-space edges, depth supervision becomes overly aggressive in smooth regions and fails to preserve structural boundaries. This leads to excessive smoothing near object edges, as depth values are inaccurately propagated across semantic boundaries. In Fig.~\ref{fig:ablation}, this is reflected in the blurred contours of the bicycle and reduced sharpness around object boundaries. The regularization provided by Canny-based edge priors proves essential in preserving high-frequency geometric features while avoiding depth overshoot in ambiguous regions.

\paragraph{\textbf{RGB-Guided Edge-Preserving Total Variation Regularization}}
The absence of $\mathcal{L}_{\text{tv}}$ results in subtle but noticeable quality degradation. Without this regularization, small texture-level inconsistencies emerge, especially in homogeneous surfaces. These inconsistencies are often manifested as noise or minor ringing artifacts in smooth regions of the rendered images. While the overall structural alignment remains mostly intact, the visual cleanliness and perceptual smoothness decrease. As seen in Fig.~\ref{fig:ablation}, the variant without $\mathcal{L}_{\text{tv}}$ introduces slight visual clutter in flat background areas, leading to lower LPIPS and marginal PSNR drops.
\begin{table}[ht]
  \centering
  \caption{Quantitative results of our model across different $Patch~Size$ settings.}
  \begin{adjustbox}{width=3.45in,center}
  \begin{tabular}{ccccccc}
    \toprule
    Dataset & $Patch~Size$ & PSNR $\uparrow$ & SSIM $\uparrow$ & LPIPS $\downarrow$ \\
    \midrule
           & 0 & 27.71 & 0.823 & 0.205 \\
    Mip-NeRF 360 & 5 & 28.21 & 0.833 & 0.181 \\
           & 10 & 28.18 & 0.830 & 0.183 \\
           & 20 & 28.20 & 0.832 & 0.185\\
    \midrule
           & 0 & 23.23 & 0.843 & 0.181 \\
       Tanks\&Temples & 5 & 23.84 & 0.852 & 0.172\\
           & 10 & 23.86 & 0.854 & 0.175 \\
           & 20 & 23.88 & 0.856 & 0.174 \\
    \bottomrule
  \end{tabular}
  \end{adjustbox}
  \label{tab:metrics}
\end{table}
\subsection{Discussion}
\textbf{Monocular Depth Estimator.} For monocular depth prediction, we employ the pre-trained Depth Anything V2 model~\cite{yang2024depth2}, which represents the current state-of-the-art in monocular depth estimation. While DPT~\cite{ranftl2021vision} has been widely used in NeRF-based pipelines, we choose Depth Anything V2 due to its superior depth prediction quality and generalization performance. While its best-performing ``giant'' variant is not publicly released, we utilize the publicly available ``large'' version (vitl), which offers a favorable trade-off between performance and accessibility. This estimator is used to generate initial depth maps that guide our regularization modules during training.

\textbf{Patch Size Selection.} In our experiments, we investigate the impact of different patch sizes on model performance. As shown in Tab~\ref{tab:metrics}, varying the patch size yields comparable results, indicating that the model is not highly sensitive to this parameter. This robustness arises from the relative formulation of our patch-based normalization, as well as the localized nature of our loss functions across multiple spatial scales. To further enhance generalization and avoid overfitting to a fixed patch resolution, we adopt a randomized patch size sampling strategy during training. Specifically, patch sizes are uniformly sampled from a predefined range ([5, 20]), which introduces diversity in spatial context and encourages stability across varying granularities.
\subsection{Limitation and Future Work}
While our method improves rendering fidelity, certain limitations remain. First, the effectiveness of depth-based supervision relies on the quality of the estimated depth maps, which may be inaccurate in texture-less or occluded regions. This can introduce biases into Gaussian placement, particularly in complex scenes. Second, although our edge-aware and TV-based regularizations enhance structural detail, they are still hand-crafted components and may not generalize optimally across all scene types or lighting conditions. In future work, we plan to explore adaptive or learned regularization strategies that can dynamically adjust to scene content. Incorporating confidence-aware depth supervision or leveraging multi-modal inputs (e.g., surface normals or segmentation cues) may further improve robustness in challenging scenarios such as reflections, transparency, or thin structures.
\section{Conclusion}
In this paper, we proposed a novel depth- and edge-guided framework for 3D Gaussian Splatting, designed to improve rendering fidelity and geometric consistency in novel view synthesis tasks. Our method introduces three key components: Hierarchical Geometric Depth Supervision for multi-scale structural alignment, Edge-Aware Depth Regularization to preserve semantic boundaries, and an RGB-Guided Edge-Preserving TV loss to suppress visual artifacts while maintaining texture details. Extensive experiments on both real-world and synthetic datasets demonstrate that our approach consistently outperforms existing state-of-the-art methods in terms of image quality and structural accuracy. Ablation studies further validate the individual contributions of each component to the overall performance. Our method effectively enhances the spatial precision of Gaussians, reduces floaters, and achieves superior multi-view consistency. This study emphasises the advantages of integrating geometric guidance into point-based rendering, representing a significant advancement towards enhanced accuracy and reliability in 3D reconstruction.

{\appendix[More Experiments Result]
In the appendix, we provide detailed per-scene evaluation results on the Mip-NeRF 360 dataset~\cite{barron2022mip}, comparing our method with baseline models. As shown, our approach consistently outperforms existing 3DGS-based methods~\cite{kerbl20233d,yu2024mip,li2024dngaussian,yang2024spec,lu2024scaffold} in terms of PSNR, SSIM~\cite{wang2004image} and LPIPS~\cite{zhang2018unreasonable} across most scenes.}

\begin{table}[H]
\centering
\caption{\textbf{PSNR comparison on the Mip-NeRF 360 dataset.}}
\begin{adjustbox}{width=3.45in,center}
\begin{tabular}{l|ccccccccc}
\hline
& bicycle           & flowers           & garden      & stump           & treehill         & room        & counter        & kitchen           & bonsai           \\
\hline
Instant-NGP & 22.17 & 20.65          & 25.07          & 23.47          & 22.37          & 29.69          & 26.69          & 29.48 & 30.69         \\
Plenoxels & 21.91 & 20.10 & 23.49          & 20.66          & 22.25          & 27.59          & 23.62          & 23.42          & 24.67          \\
Mip-NeRF360 & 24.37 & 21.73 & 26.98 & 26.40 & 22.87 & 31.63 & 29.55 & 32.23 & 33.46 \\
\hline
3D-GS       & 25.63 & 21.94 & 27.73 & 27.02 & 22.79     & 31.80 & 29.12 & 31.61 & 32.48          \\
DNGaussian & 25.69 & 21.89 & 27.75 & 27.03 & 22.90 & 31.79 & 29.15 & 31.77 & 32.88 \\
Mip-Splatting & 25.72 & 21.93 & 27.76 & 26.94 & 22.98 & 31.74 & 29.16 & 31.55 & 32.31  \\
Scaffold-GS & 25.61 & 21.74 & 27.82 & 26.79 & 23.38  & \textbf{32.14} & 29.62 & 31.81 & 32.87      \\
Spec-Gaussian & 25.89 & 21.85 & 28.07 & 27.14 & 22.57 & 32.03 & 29.92 & 32.25 & 33.34 \\
\hline
Ours & \textbf{25.95} & \textbf{21.98} & \textbf{28.09} & \textbf{27.15} & \textbf{23.41} & 32.09 & \textbf{30.07} & \textbf{32.46} & \textbf{33.40} \\
\hline
\end{tabular}
\label{tab:psnr-mip360}
\end{adjustbox}
\end{table}

\begin{table}[H]
\centering
\caption{\textbf{SSIM Comparison on the Mip-NeRF 360 dataset.}}
\begin{adjustbox}{width=3.45in,center}
\begin{tabular}{l|ccccccccc}
\hline
& bicycle           & flowers           & garden      & stump           & treehill         & room        & counter        & kitchen           & bonsai           \\
\hline
Instant-NGP & 0.512 & 0.486 & 0.701 & 0.594 & 0.542 & 0.871 & 0.817 & 0.858 & 0.906         \\
Plenoxels & 0.496 & 0.431 & 0.606 & 0.523 & 0.509 & 0.842 & 0.759 & 0.648 & 0.814          \\
Mip-NeRF360 & 0.685 & 0.583 & 0.813 & 0.744 & 0.632 & 0.913 & 0.894 & 0.920 & 0.941 \\
\hline
3D-GS       & 0.778 & 0.623 & 0.874 & 0.784 & 0.651 & 0.928 & 0.916 & 0.933 & 0.948          \\
DNGaussian & 0.779 & 0.635 & 0.877 & 0.790 & 0.654 & 0.927 & 0.917 & 0.932 & 0.944 \\
Mip-Splatting & 0.780 & 0.623 & 0.875 & 0.786 & 0.655 & 0.928 & 0.916 & 0.933 & 0.948 \\
Scaffold-GS & 0.773 & 0.609 & 0.867 & 0.774 & 0.657 & 0.931 & 0.919 & 0.931 & 0.950        \\
Spec-Gaussian & 0.796 & 0.648 & 0.881 & 0.795 & 0.645 & 0.934 & 0.922 & 0.937 & 0.952 \\
\hline
Ours & \textbf{0.801} & \textbf{0.655} & \textbf{0.882} & \textbf{0.805} & \textbf{0.661} & \textbf{0.939} & \textbf{0.924} & \textbf{0.939} & \textbf{0.953} \\
\hline
\end{tabular}
\label{tab:ssim-mip360}
\end{adjustbox}
\end{table}

\begin{table}[ht]
\centering
\caption{\textbf{LPIPS Comparison on the Mip-NeRF 360 dataset.}}
\begin{adjustbox}{width=3.45in,center}
\begin{tabular}{l|ccccccccc}
\hline
& bicycle           & flowers           & garden      & stump           & treehill         & room        & counter        & kitchen           & bonsai           \\
\hline
Instant-NGP & 0.446 & 0.441 & 0.257 & 0.421 & 0.450 & 0.261 & 0.306 & 0.195 & 0.205         \\
Plenoxels & 0.506 & 0.521 & 0.386 & 0.503 & 0.540 & 0.419 & 0.441 & 0.447 & 0.398          \\
Mip-NeRF360 & 0.301 & 0.344 & 0.170 & 0.261 & 0.339 & 0.211 & 0.204 & 0.127 & 0.176 \\
\hline
3D-GS       & 0.204 & 0.328 & 0.103 & 0.207 & 0.318 & 0.191 & 0.178 & 0.113 & 0.173          \\
DNGaussian & 0.187 & 0.321 & 0.101 & 0.206 & 0.310 & 0.186 & 0.169 & 0.110 & 0.169 \\
Mip-Splatting & 0.206 & 0.331 & 0.103 & 0.209 & 0.320 & 0.192 & 0.179 & 0.113 & 0.173\\
Scaffold-GS & 0.224 & 0.339 & 0.112 & 0.228 & 0.315 & 0.182 & 0.177 & 0.114 & 0.174          \\
Spec-Gaussian & 0.166 & 0.264 & \textbf{0.092} & 0.186 & 0.271 & 0.177 & 0.166 & 0.108 & 0.162 \\
\hline
Ours & \textbf{0.165} & \textbf{0.263} & 0.093 & \textbf{0.181} & \textbf{0.269} & \textbf{0.173} & \textbf{0.164} & \textbf{0.105} & \textbf{0.159} \\
\hline
\end{tabular}
\label{tab:lpips-mip360}
\end{adjustbox}
\end{table}

\bibliographystyle{IEEEtran}
\bibliography{IEEEabrv,references}

\begin{thebibliography}{10}
\providecommand{\url}[1]{#1}
\csname url@samestyle\endcsname
\providecommand{\newblock}{\relax}
\providecommand{\bibinfo}[2]{#2}
\providecommand{\BIBentrySTDinterwordspacing}{\spaceskip=0pt\relax}
\providecommand{\BIBentryALTinterwordstretchfactor}{4}
\providecommand{\BIBentryALTinterwordspacing}{\spaceskip=\fontdimen2\font plus
\BIBentryALTinterwordstretchfactor\fontdimen3\font minus \fontdimen4\font\relax}
\providecommand{\BIBforeignlanguage}[2]{{%
\expandafter\ifx\csname l@#1\endcsname\relax
\typeout{** WARNING: IEEEtran.bst: No hyphenation pattern has been}%
\typeout{** loaded for the language `#1'. Using the pattern for}%
\typeout{** the default language instead.}%
\else
\language=\csname l@#1\endcsname
\fi
#2}}
\providecommand{\BIBdecl}{\relax}
\BIBdecl

\bibitem{mildenhall2021nerf}
B.~Mildenhall, P.~P. Srinivasan, M.~Tancik, J.~T. Barron, R.~Ramamoorthi, and R.~Ng, ``Nerf: Representing scenes as neural radiance fields for view synthesis,'' \emph{Communications of the ACM}, vol.~65, no.~1, pp. 99--106, 2021.

\bibitem{chen2021mvsnerf}
A.~Chen, Z.~Xu, F.~Zhao, X.~Zhang, F.~Xiang, J.~Yu, and H.~Su, ``Mvsnerf: Fast generalizable radiance field reconstruction from multi-view stereo,'' in \emph{Proceedings of the IEEE/CVF international conference on computer vision}, 2021, pp. 14\,124--14\,133.

\bibitem{yang2023freenerf}
J.~Yang, M.~Pavone, and Y.~Wang, ``Freenerf: Improving few-shot neural rendering with free frequency regularization,'' in \emph{Proceedings of the IEEE/CVF conference on computer vision and pattern recognition}, 2023, pp. 8254--8263.

\bibitem{yu2021pixelnerf}
A.~Yu, V.~Ye, M.~Tancik, and A.~Kanazawa, ``pixelnerf: Neural radiance fields from one or few images,'' in \emph{Proceedings of the IEEE/CVF conference on computer vision and pattern recognition}, 2021, pp. 4578--4587.

\bibitem{muller2022instant}
T.~M{\"u}ller, A.~Evans, C.~Schied, and A.~Keller, ``Instant neural graphics primitives with a multiresolution hash encoding,'' \emph{ACM transactions on graphics (TOG)}, vol.~41, no.~4, pp. 1--15, 2022.

\bibitem{sun2022direct}
C.~Sun, M.~Sun, and H.-T. Chen, ``Direct voxel grid optimization: Super-fast convergence for radiance fields reconstruction,'' in \emph{Proceedings of the IEEE/CVF conference on computer vision and pattern recognition}, 2022, pp. 5459--5469.

\bibitem{kerbl20233d}
B.~Kerbl, G.~Kopanas, T.~Leimk{\"u}hler, and G.~Drettakis, ``3d gaussian splatting for real-time radiance field rendering.'' \emph{ACM Trans. Graph.}, vol.~42, no.~4, pp. 139--1, 2023.

\bibitem{yu2024mip}
Z.~Yu, A.~Chen, B.~Huang, T.~Sattler, and A.~Geiger, ``Mip-splatting: Alias-free 3d gaussian splatting,'' in \emph{Proceedings of the IEEE/CVF conference on computer vision and pattern recognition}, 2024, pp. 19\,447--19\,456.

\bibitem{lu2024scaffold}
T.~Lu, M.~Yu, L.~Xu, Y.~Xiangli, L.~Wang, D.~Lin, and B.~Dai, ``Scaffold-gs: Structured 3d gaussians for view-adaptive rendering,'' in \emph{Proceedings of the IEEE/CVF Conference on Computer Vision and Pattern Recognition}, 2024, pp. 20\,654--20\,664.

\bibitem{yang2024spec}
Z.~Yang, X.~Gao, Y.-T. Sun, Y.~Huang, X.~Lyu, W.~Zhou, S.~Jiao, X.~Qi, and X.~Jin, ``Spec-gaussian: Anisotropic view-dependent appearance for 3d gaussian splatting,'' \emph{Advances in Neural Information Processing Systems}, vol.~37, pp. 61\,192--61\,216, 2024.

\bibitem{10678533}
J.~Chung, J.~Oh, and K.~M. Lee, ``Depth-regularized optimization for 3d gaussian splatting in few-shot images,'' in \emph{2024 IEEE/CVF Conference on Computer Vision and Pattern Recognition Workshops (CVPRW)}, 2024, pp. 811--820.

\bibitem{yu2024lm}
H.~Yu, X.~Long, and P.~Tan, ``Lm-gaussian: Boost sparse-view 3d gaussian splatting with large model priors,'' \emph{arXiv preprint arXiv:2409.03456}, 2024.

\bibitem{shen2025dof}
L.~Shen, T.~Liu, H.~Sun, J.~Li, Z.~Cao, W.~Li, and C.~C. Loy, ``Dof-gaussian: Controllable depth-of-field for 3d gaussian splatting,'' in \emph{Proceedings of the Computer Vision and Pattern Recognition Conference}, 2025, pp. 26\,462--26\,471.

\bibitem{li2024dngaussian}
J.~Li, J.~Zhang, X.~Bai, J.~Zheng, X.~Ning, J.~Zhou, and L.~Gu, ``Dngaussian: Optimizing sparse-view 3d gaussian radiance fields with global-local depth normalization,'' in \emph{Proceedings of the IEEE/CVF conference on computer vision and pattern recognition}, 2024, pp. 20\,775--20\,785.

\bibitem{canny1986computational}
J.~Canny, ``A computational approach to edge detection,'' \emph{IEEE Transactions on pattern analysis and machine intelligence}, no.~6, pp. 679--698, 1986.

\bibitem{park2019deepsdf}
J.~J. Park, P.~Florence, J.~Straub, R.~Newcombe, and S.~Lovegrove, ``Deepsdf: Learning continuous signed distance functions for shape representation,'' in \emph{Proceedings of the IEEE/CVF conference on computer vision and pattern recognition}, 2019, pp. 165--174.

\bibitem{wang2021neus}
P.~Wang, L.~Liu, Y.~Liu, C.~Theobalt, T.~Komura, and W.~Wang, ``Neus: Learning neural implicit surfaces by volume rendering for multi-view reconstruction,'' \emph{NeurIPS}, 2021.

\bibitem{wang2023neus2}
Y.~Wang, Q.~Han, M.~Habermann, K.~Daniilidis, C.~Theobalt, and L.~Liu, ``Neus2: Fast learning of neural implicit surfaces for multi-view reconstruction,'' in \emph{Proceedings of the IEEE/CVF International Conference on Computer Vision}, 2023, pp. 3295--3306.

\bibitem{barron2021mip}
J.~T. Barron, B.~Mildenhall, M.~Tancik, P.~Hedman, R.~Martin-Brualla, and P.~P. Srinivasan, ``Mip-nerf: A multiscale representation for anti-aliasing neural radiance fields,'' in \emph{Proceedings of the IEEE/CVF international conference on computer vision}, 2021, pp. 5855--5864.

\bibitem{barron2022mip}
J.~T. Barron, B.~Mildenhall, D.~Verbin, P.~P. Srinivasan, and P.~Hedman, ``Mip-nerf 360: Unbounded anti-aliased neural radiance fields,'' in \emph{Proceedings of the IEEE/CVF conference on computer vision and pattern recognition}, 2022, pp. 5470--5479.

\bibitem{yu2021plenoctrees}
A.~Yu, R.~Li, M.~Tancik, H.~Li, R.~Ng, and A.~Kanazawa, ``Plenoctrees for real-time rendering of neural radiance fields,'' in \emph{Proceedings of the IEEE/CVF international conference on computer vision}, 2021, pp. 5752--5761.

\bibitem{liu2020neural}
L.~Liu, J.~Gu, K.~Zaw~Lin, T.-S. Chua, and C.~Theobalt, ``Neural sparse voxel fields,'' \emph{Advances in Neural Information Processing Systems}, vol.~33, pp. 15\,651--15\,663, 2020.

\bibitem{fridovich2022plenoxels}
S.~Fridovich-Keil, A.~Yu, M.~Tancik, Q.~Chen, B.~Recht, and A.~Kanazawa, ``Plenoxels: Radiance fields without neural networks,'' in \emph{Proceedings of the IEEE/CVF conference on computer vision and pattern recognition}, 2022, pp. 5501--5510.

\bibitem{chen2022tensorf}
A.~Chen, Z.~Xu, A.~Geiger, J.~Yu, and H.~Su, ``Tensorf: Tensorial radiance fields,'' in \emph{European conference on computer vision}.\hskip 1em plus 0.5em minus 0.4em\relax Springer, 2022, pp. 333--350.

\bibitem{huang2024efficient}
Z.~Huang, S.~M. Erfani, S.~Lu, and M.~Gong, ``Efficient neural implicit representation for 3d human reconstruction,'' \emph{Pattern Recognition}, vol. 156, p. 110758, 2024.

\bibitem{xu2023vr}
L.~Xu, V.~Agrawal, W.~Laney, T.~Garcia, A.~Bansal, C.~Kim, S.~Rota~Bul{\`o}, L.~Porzi, P.~Kontschieder, A.~Bo{\v{z}}i{\v{c}} \emph{et~al.}, ``Vr-nerf: High-fidelity virtualized walkable spaces,'' in \emph{SIGGRAPH Asia 2023 Conference Papers}, 2023, pp. 1--12.

\bibitem{barron2023zip}
J.~T. Barron, B.~Mildenhall, D.~Verbin, P.~P. Srinivasan, and P.~Hedman, ``Zip-nerf: Anti-aliased grid-based neural radiance fields,'' in \emph{Proceedings of the IEEE/CVF International Conference on Computer Vision}, 2023, pp. 19\,697--19\,705.

\bibitem{liao2022kitti}
Y.~Liao, J.~Xie, and A.~Geiger, ``Kitti-360: A novel dataset and benchmarks for urban scene understanding in 2d and 3d,'' \emph{IEEE Transactions on Pattern Analysis and Machine Intelligence}, vol.~45, no.~3, pp. 3292--3310, 2022.

\bibitem{schonberger2016pixelwise}
J.~L. Sch{\"o}nberger, E.~Zheng, J.-M. Frahm, and M.~Pollefeys, ``Pixelwise view selection for unstructured multi-view stereo,'' in \emph{Computer Vision--ECCV 2016: 14th European Conference, Amsterdam, The Netherlands, October 11-14, 2016, Proceedings, Part III 14}.\hskip 1em plus 0.5em minus 0.4em\relax Springer, 2016, pp. 501--518.

\bibitem{ruckert2022adop}
D.~R{\"u}ckert, L.~Franke, and M.~Stamminger, ``Adop: Approximate differentiable one-pixel point rendering,'' \emph{ACM Transactions on Graphics (ToG)}, vol.~41, no.~4, pp. 1--14, 2022.

\bibitem{kopanas2021point}
G.~Kopanas, J.~Philip, T.~Leimk{\"u}hler, and G.~Drettakis, ``Point-based neural rendering with per-view optimization,'' in \emph{Computer Graphics Forum}, vol.~40, no.~4.\hskip 1em plus 0.5em minus 0.4em\relax Wiley Online Library, 2021, pp. 29--43.

\bibitem{wiles2020synsin}
O.~Wiles, G.~Gkioxari, R.~Szeliski, and J.~Johnson, ``Synsin: End-to-end view synthesis from a single image,'' in \emph{Proceedings of the IEEE/CVF conference on computer vision and pattern recognition}, 2020, pp. 7467--7477.

\bibitem{arvanitis2021broad}
G.~Arvanitis, E.~I. Zacharaki, L.~V{\'a}{\^s}a, and K.~Moustakas, ``Broad-to-narrow registration and identification of 3d objects in partially scanned and cluttered point clouds,'' \emph{IEEE Transactions on Multimedia}, vol.~24, pp. 2230--2245, 2021.

\bibitem{huang2025structgs}
Z.~Huang, M.~Xu, and S.~Perry, ``Structgs: Adaptive spherical harmonics and rendering enhancements for superior 3d gaussian splatting,'' \emph{arXiv preprint arXiv:2503.06462}, 2025.

\bibitem{huang2025gaussianfocus}
------, ``Gaussianfocus: Constrained attention focus for 3d gaussian splatting,'' \emph{arXiv preprint arXiv:2503.17798}, 2025.

\bibitem{guo2024motion}
Z.~Guo, W.~Zhou, L.~Li, M.~Wang, and H.~Li, ``Motion-aware 3d gaussian splatting for efficient dynamic scene reconstruction,'' \emph{IEEE Transactions on Circuits and Systems for Video Technology}, 2024.

\bibitem{zhou2025gedr}
T.~Zhou, S.~Chen, S.~Wan, H.~Lv, Z.~Luo, and J.~Wu, ``Gedr: Gaussian-enhanced detail reconstruction for real-time high-fidelity 3d scene reconstruction,'' \emph{IEEE Transactions on Circuits and Systems for Video Technology}, 2025.

\bibitem{li2025frpgs}
W.~Li, X.~Pan, J.~Lin, P.~Lu, D.~Feng, and W.~Shi, ``Frpgs: Fast, robust, and photorealistic monocular dynamic scene reconstruction with deformable 3d gaussians,'' \emph{IEEE Transactions on Circuits and Systems for Video Technology}, 2025.

\bibitem{yu2024get3dgs}
H.~Yu, W.~Gong, J.~Chen, and H.~Ma, ``Get3dgs: Generate 3d gaussians based on points deformation fields,'' \emph{IEEE Transactions on Circuits and Systems for Video Technology}, 2024.

\bibitem{zheng2024gpsgaussian}
S.~Zheng, B.~Zhou, R.~Shao, B.~Liu, S.~Zhang, L.~Nie, and Y.~Liu, ``Gps-gaussian: Generalizable pixel-wise 3d gaussian splatting for real-time human novel view synthesis,'' in \emph{Proceedings of the IEEE/CVF Conference on Computer Vision and Pattern Recognition (CVPR)}, 2024.

\bibitem{qian20243dgs}
Z.~Qian, S.~Wang, M.~Mihajlovic, A.~Geiger, and S.~Tang, ``3dgs-avatar: Animatable avatars via deformable 3d gaussian splatting,'' in \emph{Proceedings of the IEEE/CVF conference on computer vision and pattern recognition}, 2024, pp. 5020--5030.

\bibitem{yan2024multi}
Z.~Yan, W.~F. Low, Y.~Chen, and G.~H. Lee, ``Multi-scale 3d gaussian splatting for anti-aliased rendering,'' in \emph{Proceedings of the IEEE/CVF Conference on Computer Vision and Pattern Recognition}, 2024, pp. 20\,923--20\,931.

\bibitem{liu2025mesh}
J.~Liu, L.~Kong, J.~Yan, and G.~Chen, ``Mesh-aligned 3d gaussian splatting for multi-resolution anti-aliasing rendering,'' \emph{IEEE Transactions on Circuits and Systems for Video Technology}, 2025.

\bibitem{wang2022uncertainty}
C.~Wang, X.~Wang, J.~Zhang, L.~Zhang, X.~Bai, X.~Ning, J.~Zhou, and E.~Hancock, ``Uncertainty estimation for stereo matching based on evidential deep learning,'' \emph{pattern recognition}, vol. 124, p. 108498, 2022.

\bibitem{wang2024robust}
X.~Wang, H.~Luo, Z.~Wang, J.~Zheng, and X.~Bai, ``Robust training for multi-view stereo networks with noisy labels,'' \emph{Displays}, vol.~81, p. 102604, 2024.

\bibitem{wang2024contrastive}
Z.~Wang, H.~Luo, X.~Wang, J.~Zheng, X.~Ning, and X.~Bai, ``A contrastive learning based unsupervised multi-view stereo with multi-stage self-training strategy,'' \emph{Displays}, vol.~83, p. 102672, 2024.

\bibitem{deng2022depth}
K.~Deng, A.~Liu, J.-Y. Zhu, and D.~Ramanan, ``Depth-supervised nerf: Fewer views and faster training for free,'' in \emph{Proceedings of the IEEE/CVF conference on computer vision and pattern recognition}, 2022, pp. 12\,882--12\,891.

\bibitem{roessle2022dense}
B.~Roessle, J.~T. Barron, B.~Mildenhall, P.~P. Srinivasan, and M.~Nie{\ss}ner, ``Dense depth priors for neural radiance fields from sparse input views,'' in \emph{Proceedings of the IEEE/CVF Conference on Computer Vision and Pattern Recognition}, 2022, pp. 12\,892--12\,901.

\bibitem{song2023darf}
J.~Song, S.~Park, H.~An, S.~Cho, M.-S. Kwak, S.~Cho, and S.~Kim, ``D{\"a}rf: Boosting radiance fields from sparse input views with monocular depth adaptation,'' \emph{Advances in Neural Information Processing Systems}, vol.~36, pp. 68\,458--68\,470, 2023.

\bibitem{wang2023sparsenerf}
G.~Wang, Z.~Chen, C.~C. Loy, and Z.~Liu, ``Sparsenerf: Distilling depth ranking for few-shot novel view synthesis,'' in \emph{Proceedings of the IEEE/CVF international conference on computer vision}, 2023, pp. 9065--9076.

\bibitem{yu2022monosdf}
Z.~Yu, S.~Peng, M.~Niemeyer, T.~Sattler, and A.~Geiger, ``Monosdf: Exploring monocular geometric cues for neural implicit surface reconstruction,'' \emph{Advances in neural information processing systems}, vol.~35, pp. 25\,018--25\,032, 2022.

\bibitem{ranftl2021vision}
R.~Ranftl, A.~Bochkovskiy, and V.~Koltun, ``Vision transformers for dense prediction,'' in \emph{Proceedings of the IEEE/CVF international conference on computer vision}, 2021, pp. 12\,179--12\,188.

\bibitem{yang2024depth1}
L.~Yang, B.~Kang, Z.~Huang, X.~Xu, J.~Feng, and H.~Zhao, ``Depth anything: Unleashing the power of large-scale unlabeled data,'' in \emph{Proceedings of the IEEE/CVF Conference on Computer Vision and Pattern Recognition}, 2024, pp. 10\,371--10\,381.

\bibitem{yang2024depth2}
L.~Yang, B.~Kang, Z.~Huang, Z.~Zhao, X.~Xu, J.~Feng, and H.~Zhao, ``Depth anything v2,'' \emph{Advances in Neural Information Processing Systems}, vol.~37, pp. 21\,875--21\,911, 2024.

\bibitem{xiong2024sparsegs}
H.~Xiong, \emph{SparseGS: Real-time 360° sparse view synthesis using Gaussian splatting}.\hskip 1em plus 0.5em minus 0.4em\relax University of California, Los Angeles, 2024.

\bibitem{zhu2024fsgs}
Z.~Zhu, Z.~Fan, Y.~Jiang, and Z.~Wang, ``Fsgs: Real-time few-shot view synthesis using gaussian splatting,'' in \emph{European conference on computer vision}.\hskip 1em plus 0.5em minus 0.4em\relax Springer, 2024, pp. 145--163.

\bibitem{schonberger2016structure}
J.~L. Schonberger and J.-M. Frahm, ``Structure-from-motion revisited,'' in \emph{Proceedings of the IEEE conference on computer vision and pattern recognition}, 2016, pp. 4104--4113.

\bibitem{knapitsch2017tanks}
A.~Knapitsch, J.~Park, Q.-Y. Zhou, and V.~Koltun, ``Tanks and temples: Benchmarking large-scale scene reconstruction,'' \emph{ACM Transactions on Graphics (ToG)}, vol.~36, no.~4, pp. 1--13, 2017.

\bibitem{hedman2018deep}
P.~Hedman, J.~Philip, T.~Price, J.-M. Frahm, G.~Drettakis, and G.~Brostow, ``Deep blending for free-viewpoint image-based rendering,'' \emph{ACM Transactions on Graphics (ToG)}, vol.~37, no.~6, pp. 1--15, 2018.

\bibitem{hu2023tri}
W.~Hu, Y.~Wang, L.~Ma, B.~Yang, L.~Gao, X.~Liu, and Y.~Ma, ``Tri-miprf: Tri-mip representation for efficient anti-aliasing neural radiance fields,'' in \emph{Proceedings of the IEEE/CVF International Conference on Computer Vision}, 2023, pp. 19\,774--19\,783.

\bibitem{chen2023neurbf}
Z.~Chen, Z.~Li, L.~Song, L.~Chen, J.~Yu, J.~Yuan, and Y.~Xu, ``Neurbf: A neural fields representation with adaptive radial basis functions,'' in \emph{Proceedings of the IEEE/CVF International Conference on Computer Vision}, 2023, pp. 4182--4194.

\bibitem{wang2004image}
Z.~Wang, A.~C. Bovik, H.~R. Sheikh, and E.~P. Simoncelli, ``Image quality assessment: from error visibility to structural similarity,'' \emph{IEEE transactions on image processing}, vol.~13, no.~4, pp. 600--612, 2004.

\bibitem{zhang2018unreasonable}
R.~Zhang, P.~Isola, A.~A. Efros, E.~Shechtman, and O.~Wang, ``The unreasonable effectiveness of deep features as a perceptual metric,'' in \emph{Proceedings of the IEEE conference on computer vision and pattern recognition}, 2018, pp. 586--595.

\end{thebibliography}

\vfill

\end{document}